\documentclass{article}
\usepackage{graphicx}
\usepackage{amsmath, amssymb}
\usepackage{xcolor}
\usepackage{hyperref}
\usepackage{geometry}
\title{Validating Vision Transformers for Otoscopy: Performance and Data-Leakage Effects}
\author{
James Ndubuisi\\
\textit{Heriot-Watt University}
\and
Fernando Auat\\
\textit{Harper University}
\and
Marta Vallejo\\
\textit{Heriot-Watt University}
}

\date{}

\begin{document}

\maketitle

\begin{abstract}
This study evaluates the efficacy of vision transformer models, specifically Swin transformers, in enhancing the diagnostic accuracy of ear diseases compared to traditional convolutional neural networks. With a reported 27\% misdiagnosis rate among specialist otolaryngologists, improving diagnostic accuracy is crucial. The research utilised a real-world dataset from the Department of Otolaryngology at the Clinical Hospital of the Universidad de Chile, comprising otoscopic videos of ear examinations depicting various middle and external ear conditions. Frames were selected based on the Laplacian and Shannon entropy thresholds, with blank frames removed. Initially, Swin v1 and Swin v2 transformer models achieved accuracies of 100\% and 99.1\%, respectively, marginally outperforming the ResNet model (99.5\%). These results surpassed metrics reported in related studies. However, the evaluation uncovered a critical data leakage issue in the preprocessing step, affecting both this study and related research using the same raw dataset. After mitigating the data leakage, model performance decreased significantly. Corrected accuracies were 83\% for both Swin v1 and Swin v2, and 82\% for the ResNet model. This finding highlights the importance of rigorous data handling in machine learning studies, especially in medical applications. The findings indicate that while vision transformers show promise, it is essential to find an optimal balance between the benefits of advanced model architectures and those derived from effective data preprocessing. This balance is key to developing a reliable machine learning model for diagnosing ear diseases.

\end{abstract}

\section{Introduction}
Hearing loss and ear-related diseases represent a significant global health challenge, affecting an estimated 1.5 billion people worldwide. The World Health Organisation (WHO) projects that by 2050, over 700 million individuals could suffer from disabling hearing loss~\cite{1}. This widespread prevalence spans all age groups, from children highly susceptible to ear infections to older adults grappling with chronic hearing impairments. In the United States alone, hearing loss affects at least 29 million people, with prevalence rates ranging from 20.6\% in adults aged 48-59 to 90\% in those over 80~\cite{2}.

Despite the critical nature of accurate diagnosis for effective treatment, current diagnostic methods for ear diseases face significant limitations. Otoscopic examinations, while routinely conducted by otolaryngologists and general practitioners, are subject to inconsistencies based on physician expertise and potential observer bias~\cite{3}. Studies have shown that non-specialists like pediatricians achieve an accurate diagnosis rate of only 50\%, while even otolaryngologists reach just 73\% accuracy~\cite{4}. These inaccuracies can lead to inappropriate treatments, resulting in severe consequences across physical, psychological, and financial domains.

Recent advancements in medical imaging and machine learning have shown promise in improving diagnostic accuracy across various medical fields~\cite{3}. However, the application of cutting-edge techniques, particularly vision transformers (ViTs), remains largely unexplored in otology. Previous attempts to automate ear infection diagnosis have primarily focused on conventional machine learning models or convolutional neural networks (CNNs), which may not fully capture the spatial dependencies present in otoscopic images~\cite{3,4,5}.

Vision transformers have emerged as a powerful tool in computer vision tasks, often outperforming CNNs by processing images in a more flexible global context~\cite{6,7}. While ViTs have demonstrated success in other medical imaging contexts such as radiology and ophthalmology~\cite{7,8}, there is a notable gap in their application to otological diagnosis.

This study aims to address this research gap by exploring the potential of Swin Transformers, a state-of-the-art vision transformer architecture, in diagnosing outer and middle ear diseases through the analysis of otoscopic videos. We hypothesise that Swin Transformer-based models will outperform conventional convolutional models in terms of accuracy and overall performance.

This study's objectives are threefold: (1) improve diagnostic accuracy in ear disease identification, (2) assess the performance of ViTs in comparison to existing machine learning approaches in otology, and (3) develop a model that can assist medical practitioners in making more informed diagnostic decisions. We aim to contribute to a transformative solution to a prevalent healthcare challenge by using the capabilities of ViTs, ultimately improving patient outcomes and reducing the global burden of ear diseases.

\section{Methods}
\subsection{Data Acquisition}
Clinicians at the Department of Otolaryngology of the Clinical Hospital of the Universidad de Chile collected the data used in this project. The protocol for the data collection with Protocol number 65 (996/18) was reviewed and approved by the Research Ethics Committee of the Hospital, procedures were conducted in line with this protocol, national regulations and the Declaration of Helsinki, revision 2013~\cite{5}. Participants of varying age groups were recruited after consulting the hospital's outpatient clinic. An otolaryngologist examined these participants using a digital otoscope DE500 firefly camera, and a raw dataset was provided as a 73GB folder with subfolders indicating the year and month of the data collected and Excel files within the folder containing diagnosis information. Due to uneven video distribution across ear conditions, this project focused on the diagnosis of four ear conditions, namely normal, chronic otitis media, ear wax and Myringosclerosis.

\begin{figure}[h!]
    \centering
    \begin{tabular}{cc}
        \includegraphics[width=0.8\textwidth,height=5cm, keepaspectratio=true]{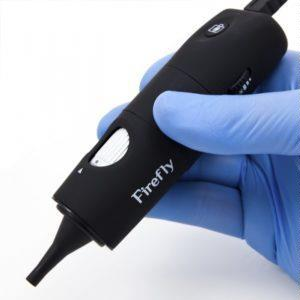}  \\
         DE500 firefly camera 
    \end{tabular}
    \caption{Firefly camera used for the examination~\cite{10}.}
    \label{fig:firefly_camera}
\end{figure}

The firefly camera records videos at $1280\times1024$ resolution and 30fps and has LEDs with adjustable brightness that illuminates the ear.

\subsection{Video Processing}
The video dataset was processed automatically. This was a multistep processing, the first step involved splitting the videos into individual frames and eliminate the initial and final frames of each video. Generally, otolaryngologists started capturing before entering the ear canal and stopped the camera after leaving the ear canal. These leading and trailing frames typically contain minimal, or no information related to the ear condition, as they often depict the camera's entry or exit from the ear canal, which is frequently obscured by motion blur. Despite this preliminary approach, the issue of blurriness persisted within the remaining frames, necessitating further processing to minimise the inclusion of frames that were either blurred or offered limited informational value.

To tackle the blurriness challenge, the Laplacian and the Shannon entropy values of each frame were used to filter out frames of the videos with low Laplacian~\cite{9} and Shannon entropy values. This translates to removing frames with less information and fewer edges. The Laplacian variance method identifies blurry images by measuring the smoothness of intensity changes. Regions with abrupt changes (edges) indicate a sharper picture, while in the context of images, Shannon entropy can be used to quantify the amount of information or randomness in the pixel values. High Shannon entropy values indicate a high degree of randomness or complexity in the image.

\begin{figure}[h!]
    \centering
    \begin{tabular}{cc}
        \includegraphics[width=0.4\textwidth,height=5cm, keepaspectratio=true]{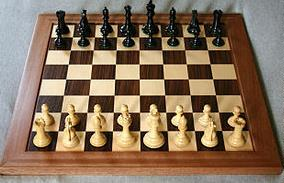} & \includegraphics[width=0.4\textwidth, height=5cm, keepaspectratio=true]{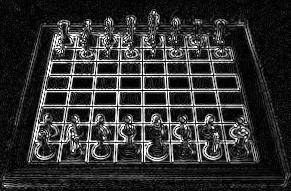} \\
        Original Image & Laplacian Image
    \end{tabular}
    \caption{Image comparison between an image of a chess board and the Laplacian image, to visually explain the Laplacian concept.}
    \label{fig:laplacian_concept}
\end{figure}

\begin{figure}[h!]
    \centering
    \includegraphics[width=0.8\textwidth,height=5cm, keepaspectratio=true]{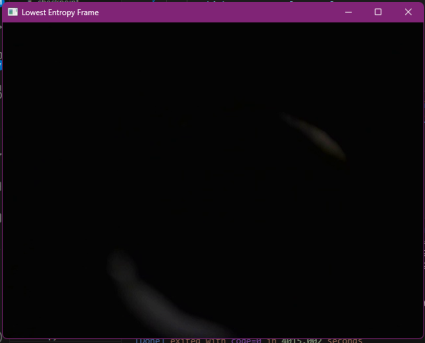}
    \includegraphics[width=0.4\textwidth, height=5cm, keepaspectratio=true]{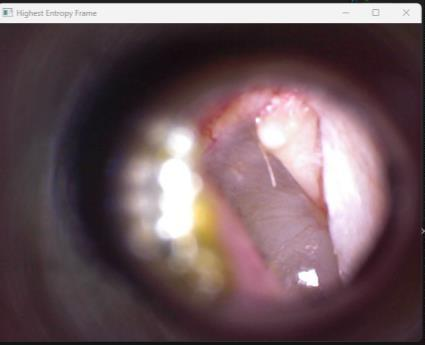}
    \caption{A side-by-side comparison of the lowest (Left) and highest (Right) entropy frames from a given video in the dataset.}
    \label{fig:entropy_frames}
\end{figure}

Due to the circular camera aperture and the nature of the images, a circular crop was applied to eliminate any information outside the region of interest.

\subsection{Model implementation}
The research utilised the TensorFlow machine learning framework with Keras for model implementation, using its simplicity and extensive deep learning resources. Three models were selected for this study: Swin Transformer V1, Swin Transformer V2, and ResNet50v2. The choice of Swin Transformers was motivated by their recent advancements in computer vision, despite their limited application in ear disease diagnosis. ResNet50v2 was included as a benchmark CNN model to evaluate the performance of the Swin Transformer models against a traditional architecture.

All models were initialised with ImageNet pretrained weights and maintained their default hyperparameters during training. (The Swin Transformer models (V1 and V2) shared common hyperparameters, including a hidden layer channel number of 96, window size of 7, and patch size of $4\times4$, among others. The ResNet50 model consisted of 48 convolutional layers, one MaxPool layer, and one Average pool layer). For all models, the learning rate was set to 0.001, using the Adam optimiser with a batch size of 32. This approach aimed to provide a fair comparison between the Swin Transformer architectures and the ResNet model in the context of ear disease classification.

\subsection{Training Hardware}
Training was conducted using Kaggle's cloud resources, featuring an NVIDIA Turing GPU with 16 GB GDDR6 memory, 2560 CUDA cores, and 320 Tensor cores, capable of 8.1 TFLOPS peak FP32 performance. The CPU was an Intel Xeon @ 2.00GHz with two cores, supported by 32GB of system RAM, running a GNU/Linux operating system. This hardware configuration provided the necessary computational power for training and comparing these advanced deep learning models in the context of ear disease classification.

\section{Results}
\subsection{Data Leakage}
A critical issue of data leakage was identified during the analysis of experimental data and related research. The problem stemmed from the method used to create train-test splits for machine learning models. Adjacent frames from the same video were inadvertently divided across training and test datasets, resulting in information from the training set leaking into the test set. This data leakage had significant implications, affecting not only the accuracy of the current models but also potentially invalidating results from previous research utilising the same dataset. To rectify this issue, a new data organisation strategy was implemented. Each patient's images were stored in separate folders, which were then allocated entirely to either the training or testing dataset. This approach effectively prevented data leakage between the sets. Additionally, an investigation was conducted to quantify the extent to which this data leakage may have inflated or impacted previously obtained results. The following metrics were used to evaluate the model: accuracy, recall, precision, Area Under the Curve (AUC), F1-score, training time, and Matthews Correlation Coefficient (MCC). The results were obtained by shuffling the dataset, splitting it, training the model on the training set, and validating it on the validation set. This was done eleven times, and the mean and variance values were recorded.

\begin{figure}[h!]
    \centering
    \begin{tabular}{cc}
        \includegraphics[width=0.45\textwidth]{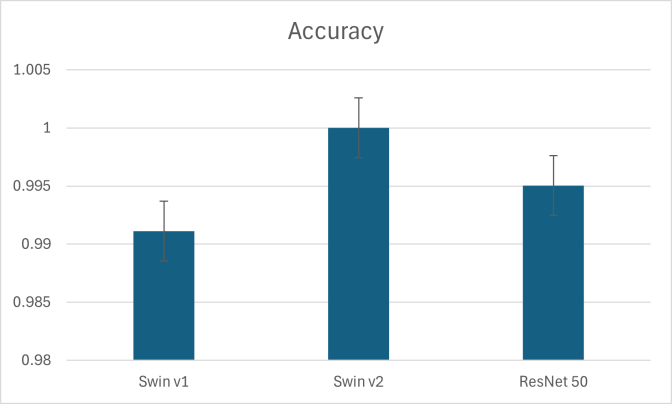} & \includegraphics[width=0.45\textwidth]{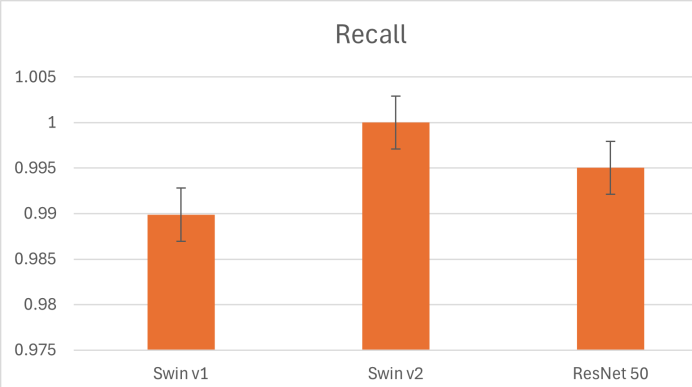}
    \end{tabular}
    \caption{Accuracy and Recall gathered from the Swin v1, Swin v2 and ResNet 50 models before addressing data leakage.}
    \label{fig:acc_recall_before}
\end{figure}

\begin{figure}[h!]
    \centering
    \includegraphics[width=0.6\textwidth]{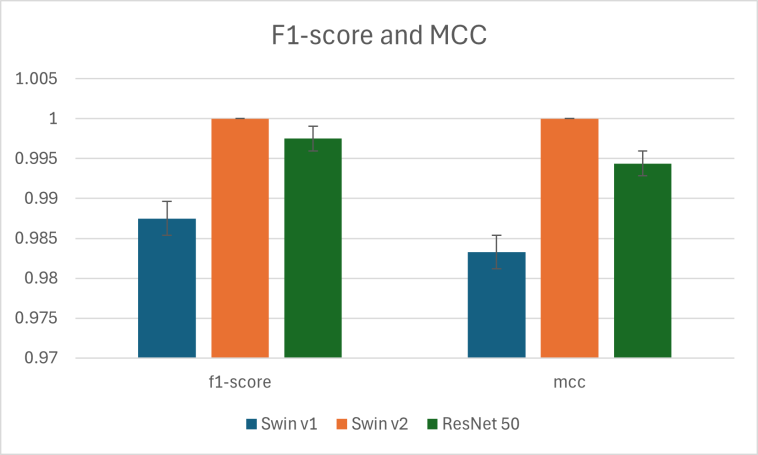}
    \caption{F1-score and MCC gathered from the Swin v1, Swin v2 and ResNet 50 models before addressing data leakage.}
    \label{fig:f1_mcc_before}
\end{figure}

These results show that the Swin v2 model had a marginally higher performance compared to the Swin v1 and ResNet 50 models. The ResNet 50 model's performance as the second-best architecture was an unanticipated outcome, deviating from initial expectations. These models deal with health data; therefore, additional evaluation metrics were employed to assess the model's classification performance across individual classes and the overall dataset.

\begin{table}[h!]
\centering
\caption{Swin v1, Swin v2 and ResNet 50 performance metrics before addressing data leakage.}
\label{tab:perf_before}
\begin{tabular}{|l|l|l|l|l|l|l|}
\hline
\textbf{Model} & \textbf{Accuracy} & \textbf{Recall} & \textbf{Precision} & \textbf{AUC} & \textbf{F1-score} & \textbf{MCC} \\ \hline
Swin v1 & 0.991122 & 0.989879 & 0.992699 & 0.999925 & 0.9875 & 0.98329 \\ \hline
Swin v2 & 1 & 1 & 1 & 1 & 1 & 1 \\ \hline
ResNet 50 & 0.995028 & 0.995028 & 0.995028 & 0.996686 & 0.9975 & 0.99439 \\ \hline
\end{tabular}
\end{table}

\begin{figure}[h!]
    \centering
    \begin{tabular}{ccc}
        \includegraphics[width=0.3\textwidth]{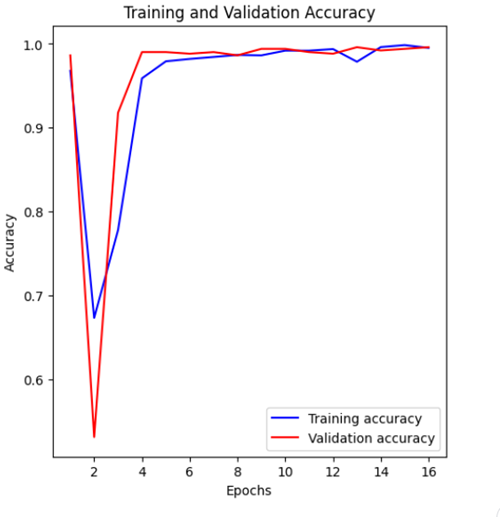} & \includegraphics[width=0.3\textwidth]{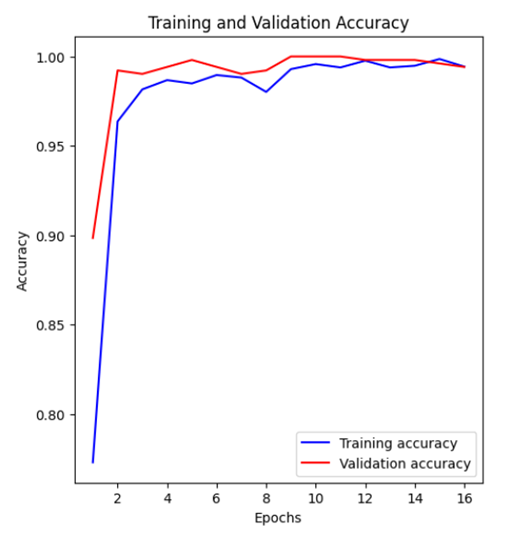} & \includegraphics[width=0.3\textwidth]{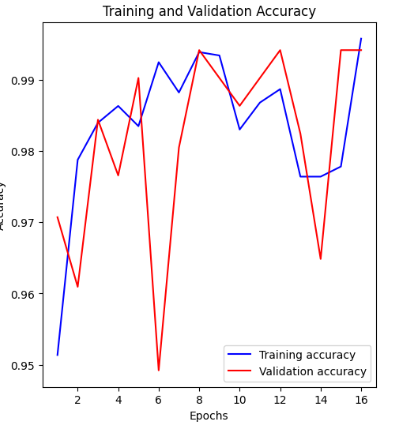} \\
        a) Swin v1 & b) Swin v2 & c) ResNet 50
    \end{tabular}
    \caption{Training Graph for Swin v1, Swin v2 and ResNet 50.}
    \label{fig:training_graphs}
\end{figure}

The Swin v2 model outperformed the rest of the models with a perfect accuracy score, closely followed by the ResNet50 model, which was a surprising result. The ResNet50 model slightly outperformed the Swin v1 model. The same trend can be seen in the recall (sensitivity) values. The graphs showing the model training process provide insight into the stability of the training process for each model. Looking at Figure~\ref{fig:training_graphs}, we can see that the training for the Swin v2 model was more consistent and stable compared to the training for the ResNet 50 and Swin v1 models. The lines on the Swin v2 training graph are smoother and more level, indicating the training progressed evenly without major fluctuations. In contrast, the graphs for ResNet 50 and Swin v1 exhibit more pronounced peaks and valleys, indicating less stable training with greater variability over time. The visual representations allow us to directly compare the relative training stability across these different model architectures.

From Figure~\ref{fig:training_graphs}, for the Swin v1 and ResNet 50 models, it is essential to select the best-performing epoch for this model to achieve a desirable performance level. Without selecting the best-performing epoch, the model's performance after 16 epochs may not be optimal.

\begin{figure}[h!]
    \centering
    \includegraphics[width=0.7\textwidth]{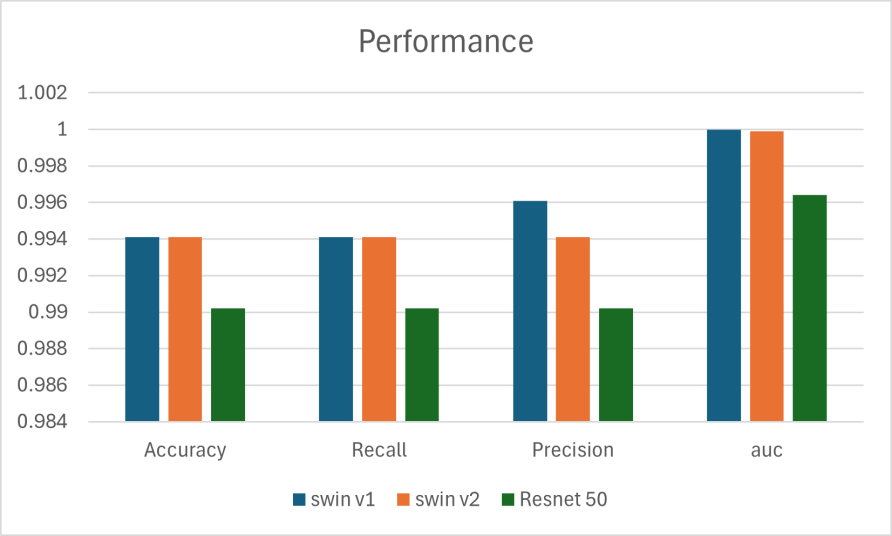}
    \caption{Performance Overview at epoch 16 before addressing data leakage.}
    \label{fig:perf_epoch16}
\end{figure}

Figure~\ref{fig:perf_epoch16} shows that the ResNet model is performing worse than the Swin transformer models when epoch 16 is selected, but this does not reflect the optimum model performance for the ResNet model within the 16 epochs. From the initial results, the ResNet50 model performed better than the Swin v1 model, but when epoch 16 is selected, the Swin v1 model now performs better than the ResNet50 model. The ResNet model accuracy was 0.4\% lower than the Swin v1 model; the same trend can be seen across the board, with the ResNet50 model performing worse than the Swin v1 model by 0.4\% in recall and AUC, 0.6\% in precision.

\begin{figure}[h!]
    \centering
    \begin{tabular}{ccc}
        \includegraphics[width=0.3\textwidth]{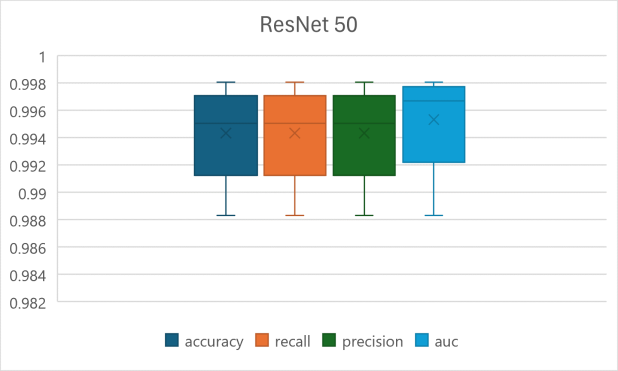} & \includegraphics[width=0.3\textwidth]{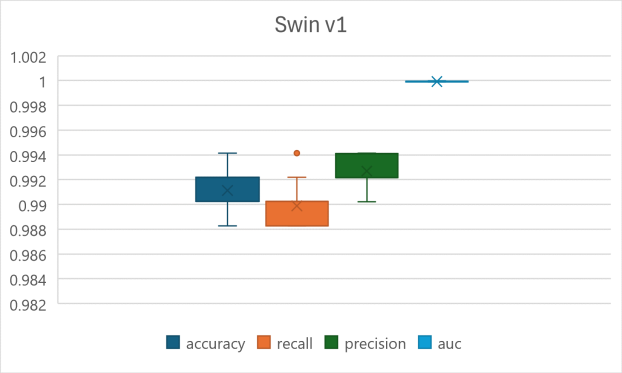} & \includegraphics[width=0.3\textwidth]{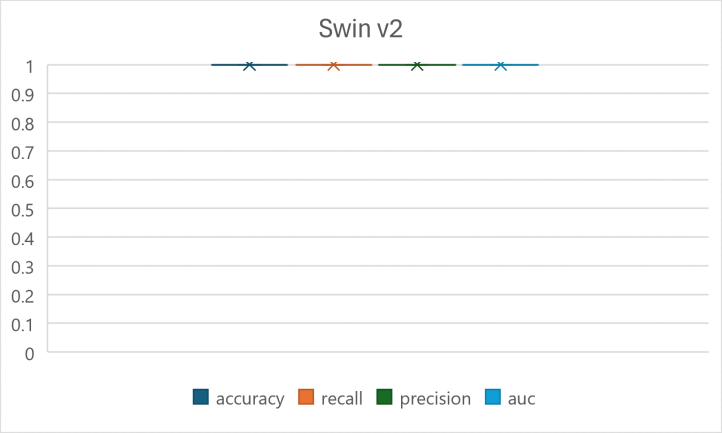} \\
        ResNet 50 & Swin v1 & Swin v2
    \end{tabular}
    \caption{Box plot of multiple runs of ResNet 50, Swin v1 and Swin v2 models on the dataset before addressing data leakage.}
    \label{fig:box_plots}
\end{figure}

To observe the model stability across multiple runs, box plots in Figure~\ref{fig:box_plots} were used to visualise the data generated by training the model 11 times.

Figure~\ref{fig:box_plots} shows that there was the most performance variance across multiple runs in the ResNet model and the least performance variance in the Swin v2 model.

\begin{figure}[h!]
    \centering
    \begin{tabular}{cc}
        \includegraphics[width=0.45\textwidth]{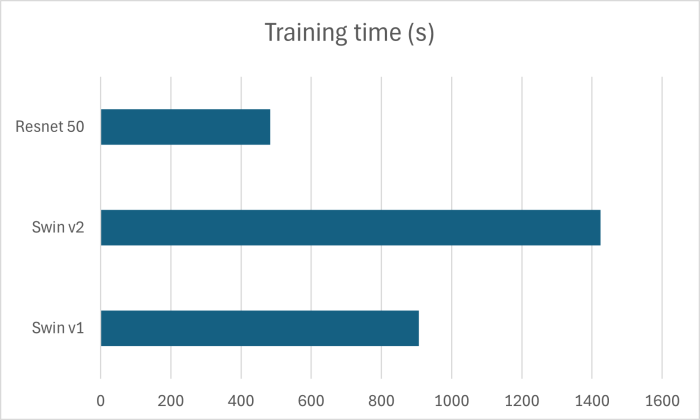} & \includegraphics[width=0.45\textwidth]{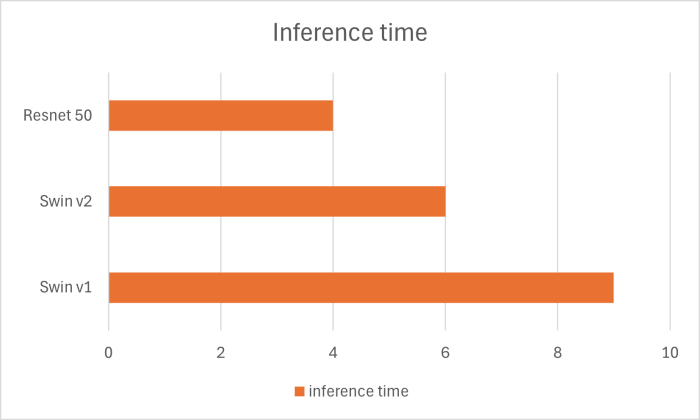}
    \end{tabular}
    \caption{Training and Inference time for Swin V1, Swin V2 and ResNet 50.}
    \label{fig:training_inference_time}
\end{figure}

A different trend emerges in terms of training time. The ResNet model requires significantly less time to achieve comparable results. After 16 epochs, the training times can be seen in Figure \ref{fig:training_inference_time}.

The ResNet model is about 2.9 times faster than the Swin v2 in terms of training time. In terms of inference time, the ResNet50 model is also twice as fast as the Swin v2 model, as seen in Figure~\ref{fig:training_inference_time}.

\begin{figure}[h!]
    \centering
    \begin{tabular}{ccc}
        \includegraphics[width=0.3\textwidth]{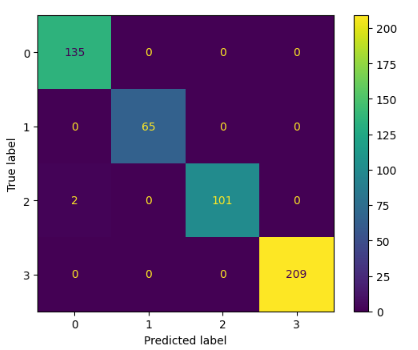} & \includegraphics[width=0.3\textwidth]{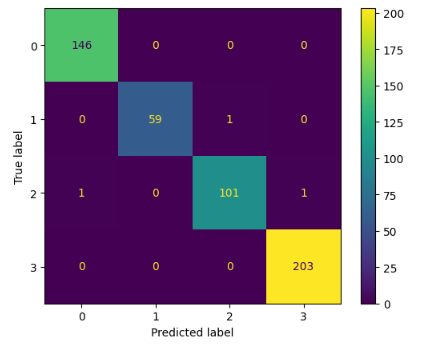} & \includegraphics[width=0.3\textwidth]{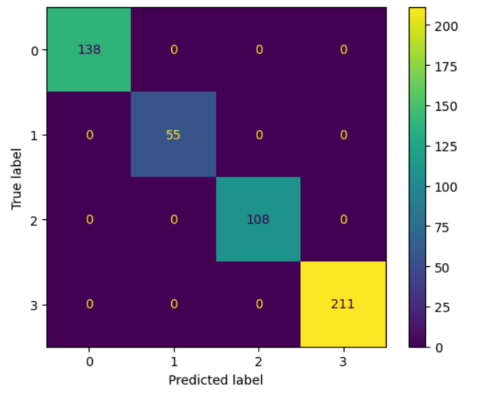} \\
        ResNet 50 & Swin v1 & Swin v2
    \end{tabular}
    \caption{Confusion Matrix for ResNet 50, Swin v1 and Swin v2 before addressing data leakage.}
    \label{fig:confusion_matrices_before}
\end{figure}

Figure~\ref{fig:confusion_matrices_before} shows the confusion matrices of the Swin v1, Swin v2 and ResNet 50 models.

In Figure~\ref{fig:confusion_matrices_before}, the confusion matrices show that Swin v2 did not misclassify any image instance, ResNet50 misclassified two images, and Swin v1 misclassified three images.

\subsubsection{Comparative Analysis}
In this section, we compare the results obtained from this study with the results obtained from existing research~\cite{3,4,5} on the same dataset. For brevity, we used the mean values from the best models across relevant research results.

\begin{table}[h!]
\centering
\caption{Comparative Analysis of Results before addressing data leakage. .}
\label{tab:comp_analysis_before}
\begin{tabular}{|l|l|l|l|l|l|}
\hline
\textbf{Model/Technique} & \textbf{Accuracy} & \textbf{Precision} & \textbf{Sensitivity} & \textbf{Specificity} & \textbf{F1-score} \\ \hline
Swin V2 & 1.00 & 1.00 & 1.00 & 1.00 & 1.00 \\ \hline
ResNet 50 & 0.995028409 & 0.995028409 & 0.995028409 & 0.996685619 & 0.9975 \\ \hline
Swin V1 & 0.991122 & 0.989879 & 0.992699 & 0.999925 & 0.9875 \\ \hline
SVM/Filter bank & 0.9903 & 0.9814 & 0.9806 & 0.9935 & * \\ \hline
ResNet 50/LSTM & 0.9784 & 0.9054 & 0.9028 & 0.9878 & 90.13 \\ \hline
VGG/Green Channel & 0.92 & 0.86 & 0.85 & 0.95 & 0.85 \\ \hline
\end{tabular}
\end{table}
{\footnotesize (*) Value not provided \\}
The Swin v2 model had the best performance compared to models from research based on the same dataset, which is closely followed by the ResNet 50 model and Swin v1 models. The models in this research outperformed those based on existing research on the same dataset. This can be attributed to the data preprocessing steps used and the novel model architectures.

\subsection{Fixed Data Leakage Results}
\begin{figure}[h!]
    \centering
    
        \includegraphics[width=0.65\textwidth]{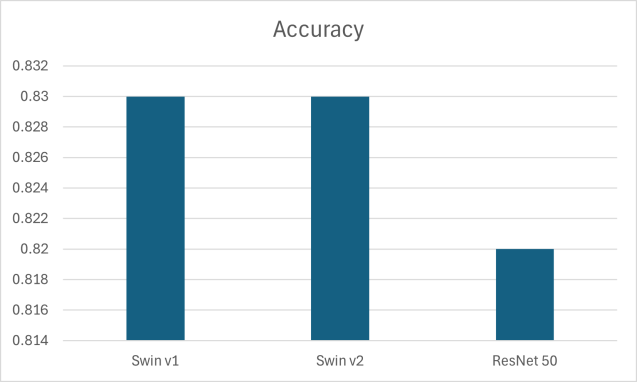}  \includegraphics[width=0.65\textwidth]{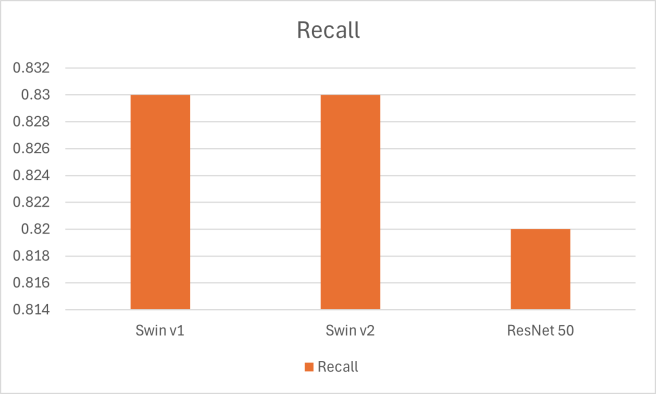}
   
    \caption{Accuracy and recall after leakage is fixed.}
    \label{fig:acc_recall_after}
\end{figure}

After fixing the data leakage by ensuring each patient image is stored in a separate folder and in either training or testing, but not both, the model was run 11 times, and the mean and standard deviation values were recorded; the results in Figure~\ref{fig:acc_recall_after} were obtained.

\begin{figure}[h!]
    \centering
    \includegraphics[width=0.7\textwidth]{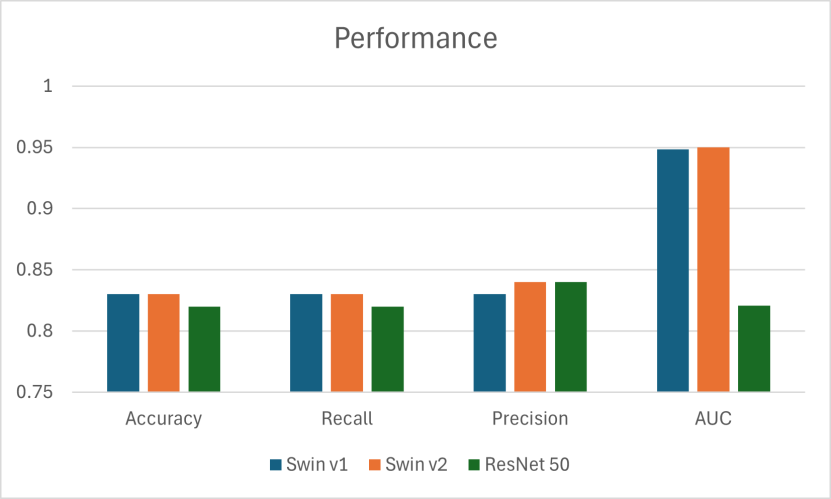}
    \caption{Performance Overview after leakage is fixed.}
    \label{fig:perf_overview_after}
\end{figure}

The accuracy of the Swin v1 and v2 models is equal, but the ResNet 50 model's accuracy is slightly lower than that of the transformer models. The same trend can be seen in the recall (sensitivity), as shown in Figure~\ref{fig:acc_recall_after}, but the values are within a 1\% range of each other. The overall model performance can be seen in Figure~\ref{fig:perf_overview_after}.

The ResNet 50 model has the same precision as the Swin v2 model, while the Swin v1 model has a slightly lower value. All the values for accuracy, precision and recall are also within 1-2\% of each other.

\begin{figure}[h!]
    \centering
    \begin{tabular}{ccc}
        \includegraphics[width=0.35\textwidth]{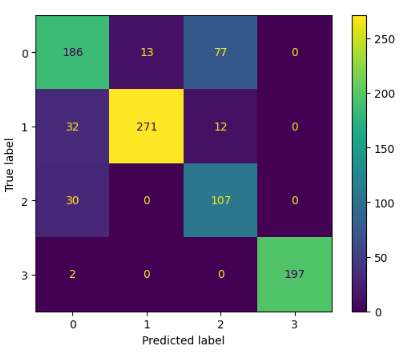} & \includegraphics[width=0.35\textwidth]{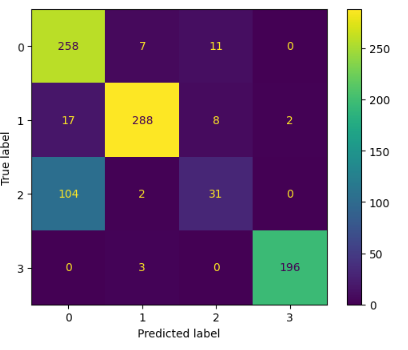} & \includegraphics[width=0.35\textwidth]{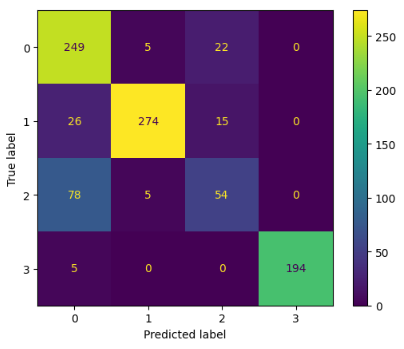} \\
        ResNet 50 & Swin v1 & Swin v2
    \end{tabular}
    \caption{Confusion matrix for ResNet 50, Swin v1 and Swin v2 models after data leakage is addressed.}
    \label{fig:confusion_matrices_after}
\end{figure}

The final confusion matrices after fixing the data leakage issue can be seen in Figure~\ref{fig:confusion_matrices_after}.

The confusion matrix in Figure~\ref{fig:confusion_matrices_after} shows that the model performs generally well across the board, but performs exceptionally well in the Normal (healthy ear class) (3) class and performs poorly in the Myringosclerosis class.

Tables \ref{tab:resnet_results_after}, \ref{tab:swinv1_results_after}, and \ref{tab:swinv2_results_after} show the accuracy, precision, recall, and F1-score of the three models across the four ear conditions.

\begin{table}[h!]
\centering
\caption{ResNet 50 results across the ear conditions after data leakage is fixed.}
\label{tab:resnet_results_after}
\begin{tabular}{|l|l|l|l|l|l|}
\hline
\textbf{Condition} & \textbf{Accuracy} & \textbf{Precision} & \textbf{Recall} & \textbf{F1-score} & \textbf{support} \\ \hline
Chronic Otitis Media & 0.74 & 0.74 & 0.67 & 0.71 & 276 \\ \hline
Earwax & 0.89 & 0.95 & 0.86 & 0.90 & 315 \\ \hline
Myringosclerosis & 0.60 & 0.55 & 0.78 & 0.64 & 137 \\ \hline
Normal & 0.99 & 1.00 & 0.99 & 0.99 & 199 \\ \hline
\end{tabular}
\end{table}

\begin{table}[h!]
\centering
\caption{Swin v1 results across the ear conditions after data leakage is fixed.}
\label{tab:swinv1_results_after}
\begin{tabular}{|l|l|l|l|l|l|}
\hline
\textbf{Condition} & \textbf{Accuracy} & \textbf{Precision} & \textbf{Recall} & \textbf{F1-score} & \textbf{support} \\ \hline
Chronic Otitis Media & 0.75 & 0.68 & 0.93 & 0.79 & 276 \\ \hline
Earwax & 0.93 & 0.96 & 0.91 & 0.94 & 315 \\ \hline
Myringosclerosis & 0.30 & 0.62 & 0.23 & 0.33 & 137 \\ \hline
Normal & 0.99 & 0.99 & 0.98 & 0.99 & 199 \\ \hline
\end{tabular}
\end{table}

\begin{table}[h!]
\centering
\caption{Swin v2 results across the ear conditions after data leakage is fixed.}
\label{tab:swinv2_results_after}
\begin{tabular}{|l|l|l|l|l|l|}
\hline
\textbf{Condition} & \textbf{Accuracy} & \textbf{Precision} & \textbf{Recall} & \textbf{F1-score} & \textbf{support} \\ \hline
Chronic Otitis Media & 0.75 & 0.68 & 0.93 & 0.79 & 276 \\ \hline
Earwax & 0.93 & 0.96 & 0.91 & 0.94 & 315 \\ \hline
Myringosclerosis & 0.30 & 0.62 & 0.23 & 0.33 & 137 \\ \hline
Normal & 0.99 & 0.99 & 0.98 & 0.99 & 199 \\ \hline
\end{tabular}
\end{table}

This poor performance on the Myringosclerosis class can be seen across all three models. This can be attributed to the limited number of training samples and the images from this class being similar enough to the chronic otitis media class that it was often misclassified as chronic otitis media.

\begin{figure}[h!]
    \centering
    \begin{tabular}{ccc}
    \includegraphics[width=0.55\textwidth]{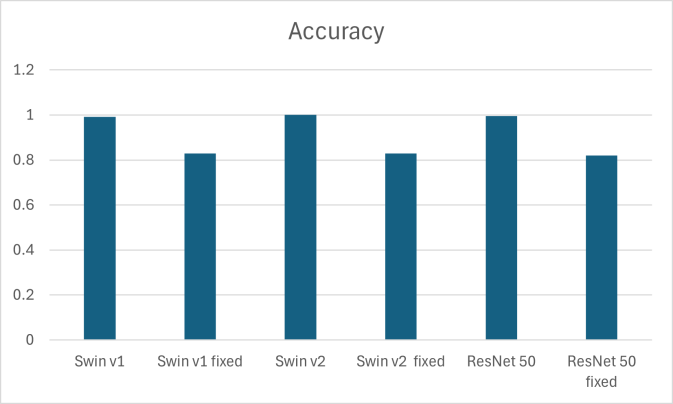}
    \includegraphics[width=0.55\textwidth]{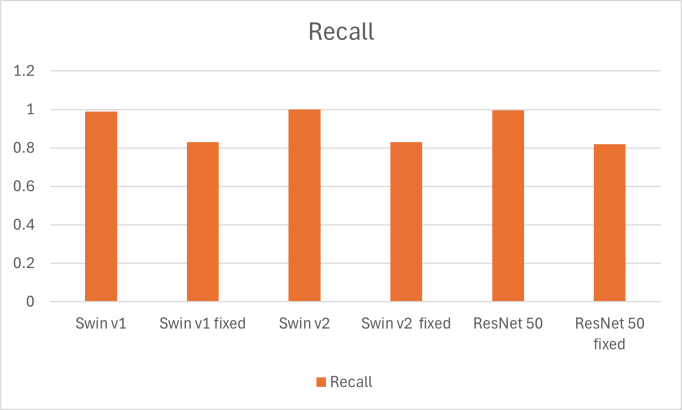}
     
    \end{tabular}
    \includegraphics[width=0.65\textwidth]{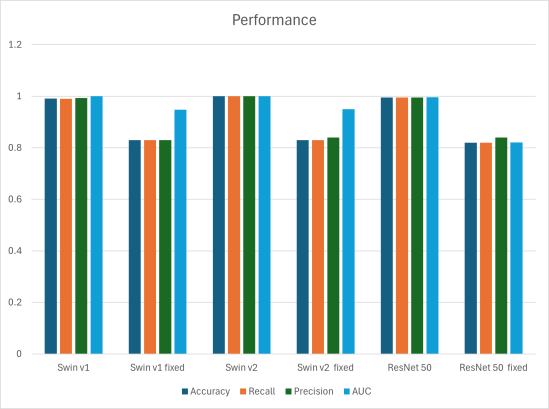}
    \caption{Comparative analysis of results before and after data leakage was fixed.}
    \label{fig:before_after_comparison}
\end{figure}

The comparison of the results obtained before and after the data leakage issue was fixed can be seen in Figure~\ref{fig:before_after_comparison}.

After addressing the data leakage issue by isolating patient data folders, we observed a substantial decline in the performance metrics across the board. On average, the accuracy, precision, recall, MCC, F1-score, and AUC values dropped by a staggering 16\%-19\%. This significant decrease in performance indicates that the data leakage had artificially inflated the performance of our models in the past.

\subsection{Anova Analysis}
In this section, we will evaluate and discuss the statistical significance of the observed drop in performance.

\begin{table}[h!]
\centering
\caption{Analysis of Variance summary}
\label{tab:anova-summary}
\begin{tabular}{|l|c|c|c|c|}
\hline
 & \textbf{Swin v1} & \textbf{Swin v2} & \textbf{ResNet 50} & \textbf{Total} \\ \hline
\multicolumn{5}{|l|}{\textit{With leakage}} \\ \hline
Count & 6 & 6 & 6 & 18 \\ \hline
Sum & 5.944414 & 6 & 5.973661 & 17.91807 \\ \hline
Mean & 0.990736 & 1 & 0.99561 & 0.995449 \\ \hline
Variance & 3.1E-05 & 0 & 1.45E-06 & 2.47E-05 \\ \hline
\multicolumn{5}{|l|}{\textit{Without leakage}} \\ \hline
Count & 6 & 6 & 6 & 18 \\ \hline
Sum & 5.0278 & 5.05298 & 4.89161 & 14.97239 \\ \hline
Mean & 0.837967 & 0.842163 & 0.815268 & 0.831799 \\ \hline
Variance & 0.003306 & 0.003373 & 0.000776 & 0.00234 \\ \hline
\multicolumn{5}{|l|}{\textbf{Total}} \\ \hline
Count & 12 & 12 & 12 & 36 \\ \hline
Sum & 10.972214 & 11.052980 & 10.865271 & 32.89046 \\ \hline
Mean & 0.914351 & 0.921082 & 0.905439 & 0.913624 \\ \hline
Variance & 0.007882 & 0.008327 & 0.009224 & *\\ \hline
Standard deviation & 0.088782 & 0.091252 & 0.096042 & *\\ \hline
\end{tabular}
\end{table}

\begin{table}[h!]
\centering
\caption{ANOVA results }

\label{tab:anova}
\begin{tabular}{|l|r|r|r|r|r|r|}
\hline
\textbf{Source of Variation} & \textbf{SS} & \textbf{df} & \textbf{MS} & \textbf{F} & \textbf{P-value} & \textbf{F crit} \\ \hline
Sample      & 0.241029 & 1  & 0.241029 & 193.1412 & 1.31E-14 & 4.170877 \\ \hline
Columns     & 0.001478 & 2  & 0.000739 & 0.592026 & 0.559539 & 3.31583  \\ \hline
Interaction & 0.001292 & 2  & 0.000646 & 0.517820 & 0.601047 & 3.31583  \\ \hline
Within      & 0.037438 & 30 & 0.001248 &    *      &    *      &    *      \\ \hline
Total       & 0.281238 & 35 &   *     & *        & *        &    *    \\ \hline
\end{tabular}
\end{table}
{\footnotesize (*) Not Applicable \\}
Table \ref{tab:anova} ANOVA results showing the Sum of Squares (SS), Degrees of Freedom (df), Mean Square (MS), F-statistic (F), Probability value (P-value), and Critical F-value (F crit).\\

In Table \ref{tab:anova}, the sample row with a p-value less than 0.05 shows a significant difference in the means between the results obtained when there was a data leakage issue in the model training and evaluation and the data obtained after the leakage was fixed. The column-row with a p-value higher than 0.05 also shows no significant difference in performance across the three models for this specific dataset.

\section{Discussion}
Initially, the analysis of experimental and research data revealed a critical issue with data leakage stemming from the storage and splitting mechanism used to create training and testing sets for the machine learning models. This resulted in adjacent frames from the same video being placed across both datasets, unintentionally leaking information from training data into the testing data. This significantly compromised the accuracy of the models, leading to unreliable and potentially misleading results. Furthermore, the issue impacted three connected research projects using the same dataset, causing inconsistencies and confusion.

A robust solution was implemented to rectify this problem. Each patient's image data was stored in separate folders, ensuring that each folder was placed exclusively in either the training or testing dataset. This approach effectively eliminated data leakage between the two sets. After addressing this leakage, a reduction in the accuracy of the machine-learning models was observed. However, the impact of the previous data leakage on the results remained to be quantified. Therefore, the next step would involve investigating how much this leakage inflated or otherwise affected the previously obtained results. This evaluation is crucial to determine the actual effectiveness of the models and the validity of the conclusions drawn from the earlier research projects. By understanding the extent of the influence of data leakage, we can ensure the reliability of future results and potentially re-evaluate the findings from the affected research.

In this study, initially, the Swin v2 model outperformed the rest of the models with a perfect accuracy score, closely followed by the ResNet50 model, which was a surprising result. The ResNet50 model slightly outperformed the Swin v1 model, and the same trend was observed in the recall (sensitivity) values. However, upon further investigation, it was discovered that the performance of the models was inflated by a data leakage issue. After addressing the issue of data leakage by properly isolating patient data folders, we observed a significant decline in performance metrics across the board, with a mean drop of a staggering 16\% to 19\% in accuracy, precision, recall, MCC, F1-score, and AUC. This significant decrease in performance exposed an uncomfortable truth – the models had been achieving misleadingly high scores by inadvertently peeking at information that would not be available in real-world applications.

Essentially, the models had been achieving misleadingly high scores by unintentionally peeking at information that would not be available in real-world applications. This peeking provided the models with an unfair advantage, enabling them to make predictions more accurately than they should have been capable of based solely on the intended input data. The realisation that the models had been achieving inflated performance due to the data leakage highlighted the importance of rigorous data isolation and validation procedures to ensure that our models are truly learning from the intended data sources and not inadvertently leveraging additional information that would not be present in real-world deployment scenarios.

However, the overall results demonstrated that the Swin v1 and Swin v2 models proved to be more reliable and consistent in terms of training stability when compared to the ResNet50 model. On the other hand, the ResNet model had significantly lower training times but exhibited high instability during training. It is crucial to select the best-performing epoch for this model to achieve a desirable performance level. Without selecting the best-performing epoch, the model's performance after 16 epochs may not be optimal.

These findings underscore the importance of carefully evaluating and selecting the appropriate model architecture and training procedures to optimise performance and ensure stability. While transformer-based models like Swin demonstrated promising results, the ResNet50 model also showed potential, albeit with higher performance variability. Ultimately, the choice of model architecture and training regimen should be informed by the specific requirements and constraints of the task at hand, considering factors such as computational resources, data characteristics, and the desired trade-offs between performance, stability, and interpretability.

\subsection{Clinical Implications}
In clinical settings, such a model could serve as a valuable decision support tool for healthcare providers, particularly in primary care or low-resource environments where access to specialist otolaryngologists is limited. By providing rapid, accurate diagnoses, the model could help reduce the current 27\% misdiagnosis rate among specialists, leading to more timely and appropriate treatments. This could potentially decrease complications from misdiagnosed or untreated ear conditions, improving patient outcomes and quality of life. Moreover, the model's ability to analyse otoscopic videos could be integrated into telemedicine platforms, enabling remote diagnosis, which is especially beneficial in rural or underserved areas. In educational contexts, the model can serve as a training tool for medical students and general practitioners, enhancing their diagnostic skills by comparing their assessments with the model's outputs. From a public health perspective, widespread implementation of such a model could facilitate large-scale screening programs, enabling early detection and intervention for ear diseases across populations. However, it is crucial to emphasise that while this model shows promise, it should be viewed as a complementary tool to clinical expertise rather than a replacement for human judgment. Future research should focus on prospective clinical trials to validate the model's performance in real-world settings and assess its impact on patient care and health outcomes.

\subsection{Limitations}
Some limitations were encountered in this research. The data labelling itself could have limited the model performance because, according to research, experts misdiagnose ear diseases 27\% of the time~\cite{4}. In some cases, ears with earwax are classified under other conditions because the earwax is not considered the official diagnosis. Secondly, the dataset was unbalanced, and some ear conditions lacked sufficient samples to provide a diverse view of the ear condition in the training and testing sets. No medical expert supervised the video processing activity to ensure that relevant frames with key information were selected.

Finally, due to the limitations in the available dataset, implementing cross-validation and ensuring no data leakage were not feasible because of the varying numbers of patients per ear condition.

\end{document}